\pdfoutput=1

\documentclass[11pt]{article}

\usepackage{EACL2023}

\usepackage{times}
\usepackage{latexsym}

\usepackage[T1]{fontenc}

\usepackage[utf8]{inputenc}

\usepackage{microtype}

\usepackage{inconsolata}

\usepackage{graphicx}
\usepackage{tabularx}
\usepackage{adjustbox}
\usepackage{booktabs}
\usepackage{multirow}
\usepackage{amsmath}
\usepackage{cuted}
\usepackage{comment}
\usepackage{lipsum}
\usepackage{colortbl}
\usepackage{xcolor}

\title{LEALLA: Learning Lightweight Language-agnostic Sentence \\ Embeddings with Knowledge Distillation}

\author{
    Zhuoyuan Mao\thanks{\ \ Currently at Kurohashi-Chu-Murawaki Lab., Kyoto University. Work done during Google internship.} \and Tetsuji Nakagawa \\
    Google Research \\
    \texttt{kevinmzy@gmail.com, tnaka@google.com} \\
}

\begin{document}
\maketitle

\begin{abstract}
Large-scale language-agnostic sentence embedding models such as LaBSE~\citep{feng-etal-2022-language} obtain state-of-the-art performance for parallel sentence alignment. However, these large-scale models can suffer from inference speed and computation overhead. This study systematically explores learning language-agnostic sentence embeddings with lightweight models. We demonstrate that a thin-deep encoder can construct robust low-dimensional sentence embeddings for 109 languages. With our proposed distillation methods, we achieve further improvements by incorporating knowledge from a teacher model. Empirical results on Tatoeba, United Nations, and BUCC show the effectiveness of our lightweight models. We release our lightweight language-agnostic sentence embedding models LEALLA on TensorFlow Hub.\footnote{\url{https://www.kaggle.com/models/google/lealla}}
\end{abstract}

\section{Introduction}
Language-agnostic sentence embedding models~\citep{artetxe-schwenk-2019-massively,yang-etal-2020-multilingual,reimers-gurevych-2020-making,feng-etal-2022-language,DBLP:journals/corr/abs-2205-15744} align multiple languages in a shared embedding space, facilitating parallel sentence alignment that extracts parallel sentences for training translation systems~\citep{schwenk-etal-2021-wikimatrix}. Among them, LaBSE~\citep{feng-etal-2022-language} achieves the state-of-the-art parallel sentence alignment accuracy over 109 languages. However, 471M parameters of LaBSE lead to the computationally-heavy inference. The 768-dimensional sentence embeddings of LaBSE (LaBSE embeddings) make it suffer from computation overhead of downstream tasks (e.g., kNN search). This limits its application on resource-constrained devices. Therefore, we explore training a lightweight model to generate low-dimensional sentence embeddings while retaining the performance of LaBSE.

We first investigate the performance of dimension-reduced LaBSE embeddings and show that it performs comparably with LaBSE. Subsequently, we experiment with various architectures to see whether such effective low-dimensional embeddings can be obtained from a lightweight encoder. We observe that the thin-deep~\citep{DBLP:journals/corr/RomeroBKCGB14} architecture is empirically superior for learning language-agnostic sentence embeddings. Diverging from previous work, we show that low-dimensional embeddings based on a lightweight model are effective for parallel sentence alignment of 109 languages.

LaBSE benefits from multilingual language model pre-training, but no multilingual pre-trained models are available for the lightweight architectures. Thus, we propose two knowledge distillation methods to further enhance the lightweight models by forcing the model to extract helpful information from LaBSE. We present three lightweight models improved with distillation: \textbf{LEALLA-small}, \textbf{LEALLA-base}, and \textbf{LEALLA-large}, with 69M, 107M, and 147M parameters, respectively. Fewer model parameters and their 128-d, 192-d, and 256-d sentence embeddings are expected to accelerate downstream tasks, while the performance drop of merely up to 3.0, 1.3, and 0.3 P@1 (or F1) points is observed on three benchmarks of parallel sentence alignment. In addition, we show the effectiveness of each loss function through an ablation study.

\section{Background: LaBSE}
LaBSE~\citep{feng-etal-2022-language} fine-tunes dual encoder models~\citep{guo-etal-2018-effective, DBLP:conf/ijcai/YangAYGSCSSK19} to learn language-agnostic embeddings from a large-scale pre-trained language model~\citep{conneau-etal-2020-unsupervised}. LaBSE is trained with parallel sentences, and each sentence pair is encoded separately by a 12-layer Transformer encoder. The 768-d encoder outputs are used to compute the training loss and serve as sentence embeddings for downstream tasks. Expressly, assume that the sentence embeddings for parallel sentences in a batch are $\{(\mathbf{x}_{i}, \mathbf{y}_{i})\}_{i=1}^{N}$ where $N$ denotes the number of the sentence pairs within a batch. LaBSE trains the bidirectional additive margin softmax (AMS) loss:
\begin{equation}
    \mathcal{L}_{ams} = \frac{1}{N} \sum_{i=1}^N(\mathcal{L}(\mathbf{x}_i,\mathbf{y}_i) + \mathcal{L}(\mathbf{y}_i,\mathbf{x}_i)),
    \label{eq:ams1}
\end{equation}
where the loss for a specific sentence pair in a single direction is defined as:
\begin{equation}
    \mathcal{L}(\mathbf{x}_i, \mathbf{y}_i)= -\log \frac{e^{\phi\left(\mathbf{x}_i, \mathbf{y}_i\right)-m}}{e^{\phi\left(\mathbf{x}_i, \mathbf{y}_i\right)-m}+\sum_{n \neq i} e^{\phi\left(\mathbf{x}_i, \mathbf{y}_n\right)}}.
    \label{eq:ams2}
\end{equation}
$m$ is a margin for optimizing the separation between translations and non-translations. $\phi\left(\mathbf{x}_i, \mathbf{y}_i\right)$ is defined as Cosine Similarity between $\mathbf{x}_i$ and $\mathbf{y}_i$.

\section{Light Language-agnostic Embeddings}
To address the efficiency issue of LaBSE, we probe the lightweight model for learning language-agnostic embeddings with the following experiments: (1) We directly reduce the dimension of LaBSE embeddings to explore the optimal embedding dimension; (2) We shrink the model size with various ways to explore the optimal architecture.

\subsection{Evaluation Settings}
\label{sec:eval}

We employ Tatoeba~\citep{artetxe-schwenk-2019-massively}, United Nations (UN)~\citep{ziemski-etal-2016-united}, and BUCC~\citep{ZWEIGENBAUM18.12} benchmarks for evaluation, which assess the model performance for parallel sentence alignment. Following~\citet{feng-etal-2022-language} and~\citet{artetxe-schwenk-2019-massively}, we report the average P@1 of bidirectional retrievals for all the languages of Tatoeba, the average P@1 for four languages of UN, and the average F1 of bidirectional retrievals for four languages of BUCC.\footnote{For BUCC, we use margin-based scoring~\citep{artetxe-schwenk-2019-margin} for filtering translation pairs.} Refer to Appx.~\ref{sec:app-eval} for details.

\subsection{Exploring the Optimal Dimension of Language-agnostic Sentence Embeddings}
\label{sec:labse-rd}

\begin{figure}[t]
    \centering
    \includegraphics[width=0.87\linewidth]{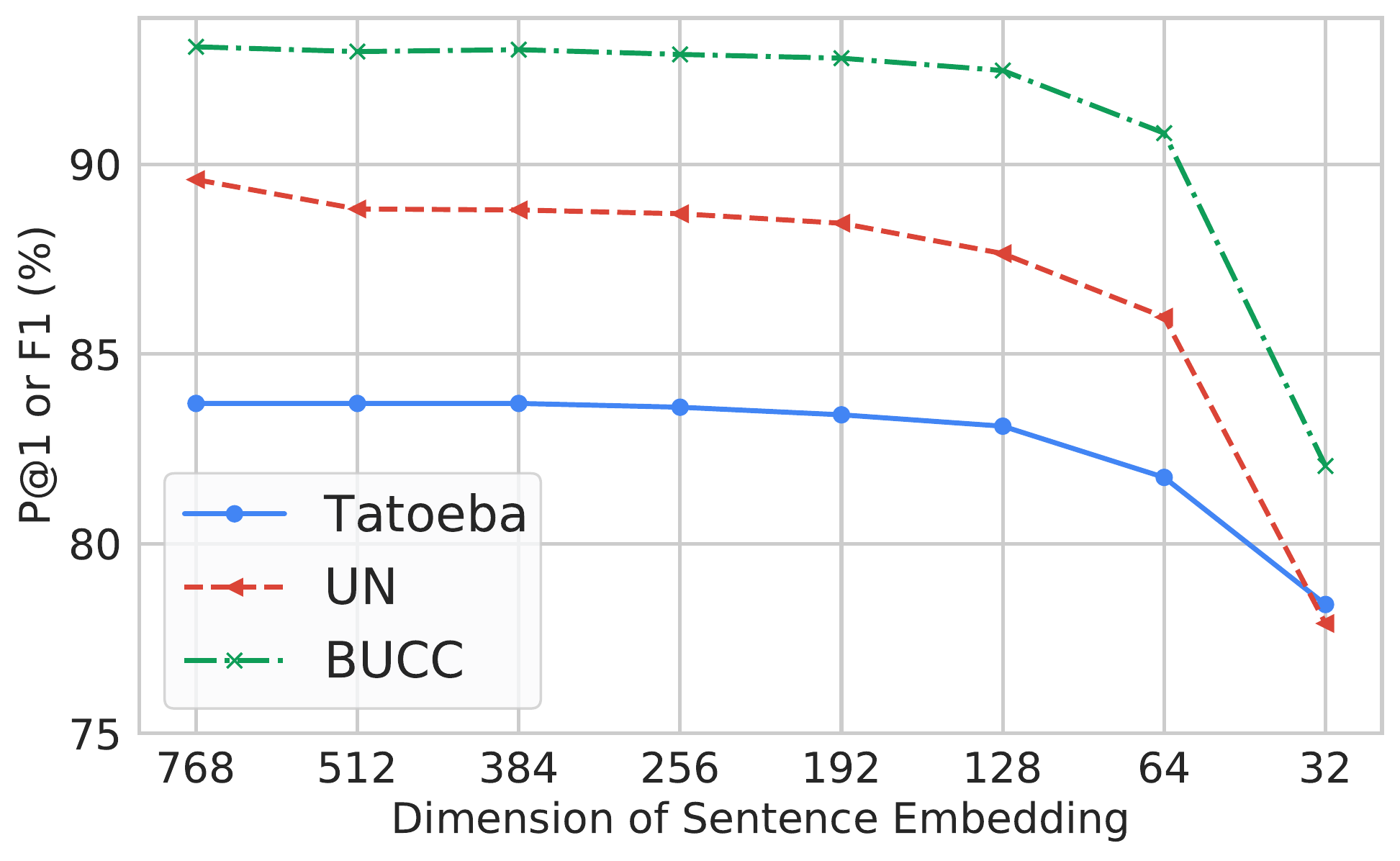}
    \caption{Dimension reduction for LaBSE.}
    \label{fig:labse-to-x}
\end{figure}

\citet{mao-etal-2021-lightweight} showed that a 256-d bilingual embedding space could achieve an accuracy of about 90\% for parallel sentence alignment. However, existing multilingual sentence embedding models such as LASER~\citeyearpar{artetxe-schwenk-2019-massively}, SBERT~\citeyearpar{reimers-gurevych-2020-making}, EMS~\citeyearpar{DBLP:journals/corr/abs-2205-15744}, and LaBSE use 768-d or 1024-d sentence embeddings, and whether a low-dimensional space can align parallel sentences over tens of languages with a solid accuracy ($>$80\%) remains unknown. Thus, we start with the dimension reduction experiments for LaBSE to explore the optimal dimension of language-agnostic sentence embeddings.

We add an extra dense layer on top of LaBSE to transform the dimension of LaBSE embeddings from 768 to lower values. We experiment with seven lower dimensions ranging from 512 to 32. We fine-tune 5k steps for fitting the newly added dense layer, whereas other parameters of LaBSE are fixed. Refer to Appx.~\ref{sec:app-training} for training details.

As shown in Fig.~\ref{fig:labse-to-x}, the performance drops more than 5 points when the dimension is 32 on Tatoeba, UN, and BUCC. Meanwhile, given sentence embeddings with a dimension over 128, they performs slightly worse than 768-d LaBSE embeddings with a performance drop of fewer than 2 points, showing that low-dimensional sentence embeddings can align parallel sentences in multiple languages. Refer to Appx.~\ref{app:labse-to-x} for detailed results.


\subsection{Exploring the Optimal Architecture}
\label{sec:lite-arch}

Although we revealed the effectiveness of the low-dimensional embeddings above, it is generated from LaBSE with 471M parameters. Thus, we explore whether such low-dimensional sentence embeddings can be obtained from an encoder with less parameters. We first reduce the number of layers (\#1 and \#2 in Table~\ref{tab:lightweight}) and the size of hidden states (\#3 and \#4) to observe the performance. Subsequently, inspired by the effectiveness of FitNet~\citep{DBLP:journals/corr/RomeroBKCGB14} and MobileBERT~\citep{sun-etal-2020-mobilebert} and taking advantage of the low-dimensional sentence embeddings shown above, we experiment with thin-deep architectures with 24 layers (\#5 - \#8), leading to fewer encoder parameters.\footnote{Following MobileBERT, we attempted architectures that have an identical size for hidden state and feed-forward hidden state, but it works poorly than \#5 - \#8. (Refer to Appx.~\ref{sec:app-mobile})} Refer to Appx.~\ref{sec:app-training} for training details.

\begin{table}[t]
    \centering
    \resizebox{0.95\linewidth}{!}{
    \begin{tabular}{crrrrrrrrr}
        \toprule
        \textbf{\#} & $\mathbf{L}$ & $\mathbf{d_h}$ & $\mathbf{H}$ & $\mathbf{P}$ & $\mathbf{P_E}$ & \textbf{Tatoeba} & \textbf{UN} & \textbf{BUCC} \\
        \toprule
        \multicolumn{9}{l}{\textbf{LaBSE}} \\
        0 & 12 & 768 & 12 & 471M & 85M & 83.7 & 89.6 & 93.1 \\
        \hline
        \multicolumn{9}{l}{\textbf{Fewer Layers}} \\
        1 & 6 & 768 & 12 & 428M & 42M & 82.9 & 88.6 & 91.9 \\
        2 & 3 & 768 & 12 & 407M & 21M & 82.2 & 87.5 & 91.2 \\
        \hline
        \multicolumn{9}{l}{\textbf{Smaller Hidden Size}} \\
        3 & 12 & 384 & 12 & 214M & 21M & 82.6 & 88.4 & 92.1 \\
        4 & 12 & 192 & 12 & 102M & 6M & 81.0 & 87.0 & 91.3 \\
        \hline
        \multicolumn{9}{l}{\textbf{Thin-deep Architecture}} \\
        5 & 24 & 384 & 12 & 235M & 42M & 83.2 & 88.6 & 92.4 \\
        6 & 24 & 256 & 8 & 147M & 19M & 82.9 & 88.5 & 92.2 \\
        7 & 24 & 192 & 12 & 107M & 11M & 81.7 & 87.4 & 91.9 \\
        8 & 24 & 128 & 8 & 69M & 5M & 80.3 & 86.3 & 90.4 \\
        \bottomrule
    \end{tabular}
    }
    \caption{Results of LaBSE variants. $\mathbf{L}$, $\mathbf{d_h}$, $\mathbf{H}$, $\mathbf{P}$, and $\mathbf{P_E}$ denote the number of layers, dimension of hidden states, number of attention heads, number of parameters, and number of encoder parameters (except for the word embedding layer). Refer to Appx.~\ref{sec:app-mobile} for detailed results.}
    \label{tab:lightweight}
\end{table}

We report the results in Table~\ref{tab:lightweight}. First, architectures with fewer layers (\#1 and \#2) perform worse than LaBSE on all three tasks and can only decrease parameters by less than 15\%. Second, increasing the number of layers (\#5 and \#7) improves the performance of 12-layer models (\#3 and \#4) with a limited parameter increase less than 10\%. Referring to LaBSE (\#0), low-dimensional embeddings from thin-deep architectures (\#5 - \#8) obtain solid results on three benchmarks with performance drops of only 3.4 points at most. Until this point, we showed that thin-deep architecture is effective for learning language-agnostic sentence embeddings.

\section{Knowledge Distillation from LaBSE}
\label{sec:kd}

Besides the large model capacity, multilingual language model pre-training benefits LaBSE for parallel sentence alignment. As no multilingual pre-trained language models are available for lightweight models we investigated in Section~\ref{sec:lite-arch}, we instead explore extracting helpful knowledge from LaBSE.

\subsection{Methodology}

\begin{figure}[t]
    \centering
    \includegraphics[width=0.95\linewidth]{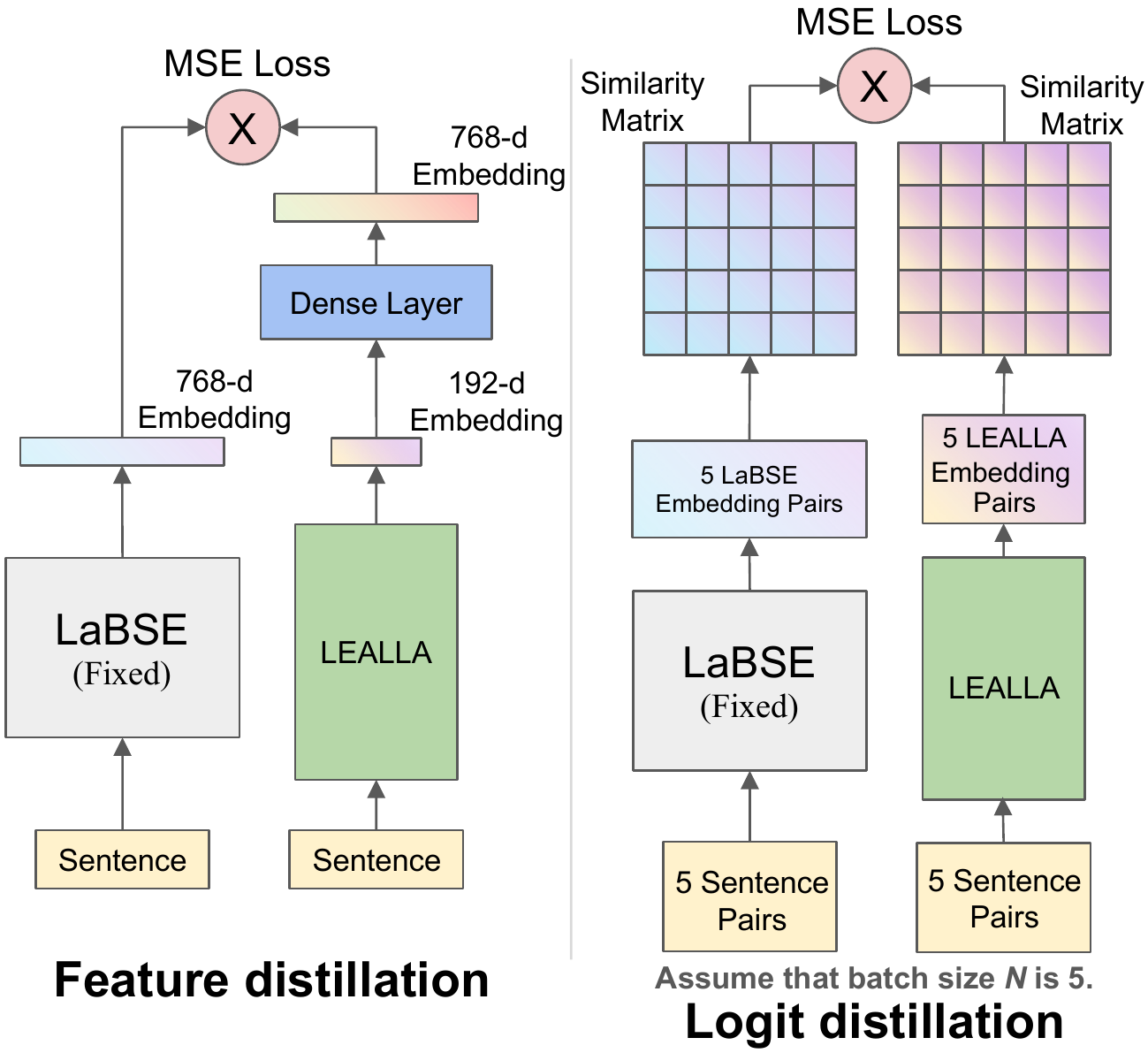}
    \caption{Feature and logit distillation from LaBSE.}
    \label{fig:distillation}
\end{figure}

\begin{table*}[t]
    \centering
    \resizebox{\linewidth}{!}{
    \begin{tabular}{l|rrr|r|rrrrr|rrrrr}
        \toprule
        \multirow{2}{*}{\textbf{Model}} & \multirow{2}{*}{$\mathbf{La.}$} & \multirow{2}{*}{$\mathbf{d}$} & \multirow{2}{*}{$\mathbf{P}$} & \multirow{2}{*}{\textbf{Ttb.}} & \multicolumn{5}{c|}{\textbf{UN}} & \multicolumn{5}{c}{\textbf{BUCC}} \\
        & & & & & \textbf{es} & \textbf{fr} & \textbf{ru} & \textbf{zh} & \textbf{avg.} & \textbf{de} & \textbf{fr} & \textbf{ru} & \textbf{zh} & \textbf{avg.} \\
        \toprule
        LASER~\citeyearpar{artetxe-schwenk-2019-massively} & 93 & 1024 & 154M & 65.5 & - & - & - & - & - & \textbf{95.4} & \textbf{92.4} & \textbf{92.3} & 91.7 & \textbf{93.0} \\
        \textit{m}-USE~\citeyearpar{yang-etal-2020-multilingual} & 16 & 512 & \textbf{85M} & - & 86.1 & 83.3 & 88.9 & 78.8 & 84.3 & 88.5 & 86.3 & 89.1 & 86.9 & 87.7 \\
        SBERT~\citeyearpar{reimers-gurevych-2020-making} & 50 & 768 & 270M & 67.1 & - & - & - & - & - & 90.8 & 87.1 & 88.6 & 87.8 & 88.6 \\
        EMS~\citeyearpar{DBLP:journals/corr/abs-2205-15744} & 62 & 1024 & 148M & 69.2 & - & - & - & - & - & 93.3 & 90.2 & 91.3 & \textbf{92.1} & 91.7 \\
        LaBSE~\citeyearpar{feng-etal-2022-language} & \textbf{109} & 768 & 471M & \textbf{83.7} & \textbf{90.8} & \textbf{89.0} & \textbf{90.4} & \textbf{88.3} & \textbf{89.6} & \textbf{95.5} & \textbf{92.3} & \textbf{92.2} & \textbf{92.5} & \textbf{93.1} \\
        \hline
        \textbf{LEALLA-small} & \textbf{109} & \textbf{128} & \textbf{69M} & 80.7 & 89.4 & 86.0 & 88.7 & 84.9 & 87.3 & 94.0 & 90.6 & 91.2 & 90.3 & 91.5 \\
        \textbf{LEALLA-base} & \textbf{109} & \textbf{192} & \textbf{107M} & \textbf{82.4} & \textbf{90.3} & \textbf{87.4} & \textbf{89.8} & \textbf{87.2} & \textbf{88.7} & 94.9 & 91.4 & 91.8 & 91.4 & 92.4 \\
        \textbf{LEALLA-large} & \textbf{109} & \textbf{256} & 147M & \textbf{83.5} & \textbf{90.8} & \textbf{88.5} & \textbf{89.9} & \textbf{87.9} & \textbf{89.3} & \textbf{95.3} & \textbf{92.0} & \textbf{92.1} & \textbf{91.9} & \textbf{92.8} \\
        \bottomrule
    \end{tabular}
    }
    \caption{Results of LEALLA. We mark the best 3 scores in \textbf{bold}. $\mathbf{La.}$, $\mathbf{d}$, $\mathbf{P}$, and \textbf{Ttb.} indicate the number of languages, dimension of sentence embeddings, number of parameters, and Tatoeba.}
    \label{tab:LEALLA}
\end{table*}

Feature distillation and logit distillation have been proven to be effective paradigms for knowledge distillation~\citep{DBLP:journals/corr/HintonVD15,DBLP:journals/corr/RomeroBKCGB14,DBLP:conf/cvpr/YimJBK17,DBLP:journals/corr/abs-1903-12136}. In this section, we propose methods applying both paradigms to language-agnostic sentence embedding distillation. We use LaBSE as a teacher to train students with thin-deep architectures which were discussed in Section~\ref{sec:lite-arch}.

\noindent\textbf{Feature Distillation}
We propose applying feature distillation to language-agnostic sentence embedding distillation, which enables lightweight sentence embeddings to approximate the LaBSE embeddings via an extra dense layer. We employ an extra trainable dense layer on top of the lightweight models to unify the embedding dimension of LaBSE and lightweight models to be 768-d, as illustrated in Fig.~\ref{fig:distillation}.\footnote{SBERT~\citeyearpar{reimers-gurevych-2020-making} used feature distillation to make monolingual sentence embeddings multilingual, but distillation between different embedding dimensions has not been studied.}\footnote{We investigated another two patterns to unify the embedding dimensions in Appx.~\ref{app:distillation}, but they performed worse.} The loss function is defined as follows:
\begin{equation}
\resizebox{\linewidth}{!}{
$
    \mathcal{L}_{fd} = \frac{1}{N} \sum_{i=1}^N(\parallel\mathbf{x}_{i}^{t} - f(\mathbf{x}_{i}^{s})\parallel_2^2+\parallel\mathbf{y}_{i}^{t} - f(\mathbf{y}_{i}^{s})\parallel_2^2),
    \label{eq:fd}
$
}
\end{equation}
where $\mathbf{x}^{t}$ (or $\mathbf{y}^{t}$) and $\mathbf{x}^{s}$ (or $\mathbf{y}^{s}$) are the embeddings by LaBSE and the lightweight model, respectively. $f(\cdot)$ is a trainable dense layer transforming the dimension from $d$ ($d<768$) to $768$.

\noindent\textbf{Logit Distillation}
We also propose applying logit distillation to language-agnostic sentence embedding distillation to extract knowledge from the sentence similarity matrix as shown in Fig.~\ref{fig:distillation}. Logit distillation forces the student to establish similar similarity relationships between the given sentence pairs as the teacher does. We propose the following mean squared error (MSE) loss:
\begin{equation}
\resizebox{\linewidth}{!}{
$
    \mathcal{L}_{ld} = \frac{1}{N^2} \sum_{i=1}^N\sum_{j=1}^N\left(\left(\phi\left(\mathbf{x}_{i}^{t}, \mathbf{y}_{j}^{t}\right) - \phi\left(\mathbf{x}_{i}^{s}, \mathbf{y}_{j}^{s}\right)\right) / T\right)^2,
    \label{eq:ld}
$
}
\end{equation}
where $T$ is a distillation temperature, and other notations follow those in Eq.~\ref{eq:ams2} and~\ref{eq:fd}.

\noindent\textbf{Combined Loss}
Finally, we combine two knowledge distillation loss functions with the AMS loss (Eq.~\ref{eq:ams1}) to jointly train the lightweight model:
\begin{equation}
    \mathcal{L}_{lealla} = \alpha\mathcal{L}_{ams} + \beta\mathcal{L}_{fd} + \gamma\mathcal{L}_{ld}.
    \label{eq:loss}
\end{equation}
Here $\alpha$, $\beta$, and $\gamma$ are weight hyperparameters, which are tuned with the development data.

\begin{figure}[t]
    \centering
    \includegraphics[width=0.9\linewidth]{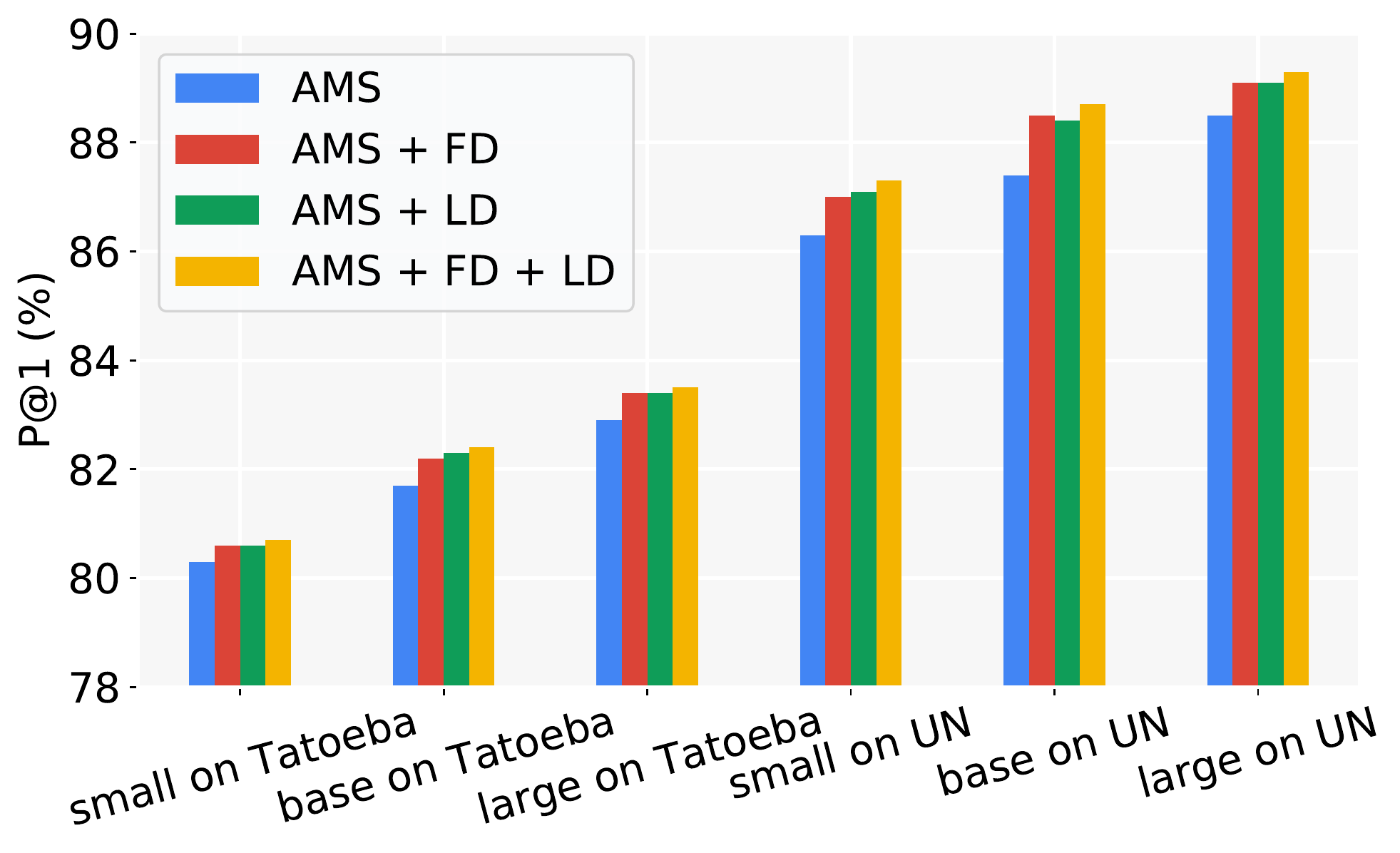}
    \caption{LEALLA with different loss combinations. AMS, FD, and LD mean $\mathcal{L}_{ams}$, $\mathcal{L}_{fd}$, and $\mathcal{L}_{ld}$.}
    \label{fig:ablation}
\end{figure}

\subsection{Experiments}
\label{sec:exps}

\noindent\textbf{Training}
We train three models, \textbf{LEALLA-small}, \textbf{LEALLA-base}, and \textbf{LEALLA-large}, using the thin-deep architectures of \#8, \#7, and \#6 in Table~\ref{tab:lightweight} and the training loss of Eq.~\ref{eq:loss}. Refer to Appx.~\ref{sec:app-training} for training and hyperparameter details.

\noindent\textbf{Results}
The results of LEALLA on Tatoeba, UN, and BUCC benchmarks are presented in Table~\ref{tab:LEALLA}. Overall, LEALLA can yield competitive performance compared with previous work. LEALLA-large performs comparably with LaBSE, where the average performance difference on three tasks is below 0.3 points. LEALLA-base and LEALLA-small obtain strong performance for high-resource languages on UN and BUCC, with a performance decrease less than 0.9 and 2.3 points, respectively. They also achieve solid results on Tatoeba with 1.3 and 3 points downgrades compared with LaBSE. The solid performance of LEALLA on Tatoeba demonstrates that it is effective for aligning parallel sentences for more than 109 languages. Moreover, all the LEALLA models perform better or comparably with previous studies other than LaBSE.

\begin{table}[t]
    \centering
    \resizebox{\linewidth}{!}{
    \begin{tabular}{l|rr|rr|rr}
        \toprule
        \multirow{2}{*}{\textbf{Loss}} & \multicolumn{2}{c|}{\textbf{LEALLA-small}} & \multicolumn{2}{c|}{\textbf{LEALLA-base}} & \multicolumn{2}{c}{\textbf{LEALLA-large}} \\
        & \textbf{Tatoeba} & \textbf{UN} & \textbf{Tatoeba} & \textbf{UN} & \textbf{Tatoeba} & \textbf{UN} \\
        \toprule
        \textit{all} & \textit{80.7} & \textit{87.3} & \textit{82.4} & \textit{88.7} & \textit{83.5} & \textit{89.3} \\
        \hline
        $\mathcal{L}_{ams}$ & \textbf{80.3} & \textbf{86.3} & \textbf{81.7} & 87.4 & \textbf{82.9} & \textbf{88.5} \\
        $\mathcal{L}_{fd}$ & 78.2 & 85.2 & 81.1 & \textbf{88.1} & 82.4 & 88.1 \\
        $\mathcal{L}_{ld}$ & 75.1 & 2.3 & 80.6 & 63.1 & 82.3 & 84.1 \\
        \bottomrule
    \end{tabular}
    }
    \caption{Results of LEALLA with each loss function. ``\textit{all}'' denotes LEALLA without ablation (with all the loss functions).}
    \label{tab:ablation}
\end{table}

\noindent\textbf{Ablation Study}
We inspect the effectiveness of each loss component in an ablative manner. First, we compare settings with and without distillation loss functions. As shown in Fig.~\ref{fig:ablation}, by adding $\mathcal{L}_{fd}$ or $\mathcal{L}_{ld}$, LEALLA trained only with $\mathcal{L}_{ams}$ is improved on Tatoeba and UN tasks. By further combining $\mathcal{L}_{fd}$ and $\mathcal{L}_{ld}$, LEALLA consistently achieves superior performance. Second, we separately train LEALLA with each loss. Referring to the results reported in Table~\ref{tab:ablation}, LEALLA trained only with $\mathcal{L}_{fd}$ yields solid performance in the ``small'' and ``base'' models compared with $\mathcal{L}_{ams}$, showing that distillation loss benefits parallel sentence alignment. $\mathcal{L}_{fd}$ and $\mathcal{L}_{ld}$ perform much worse in the ``small'' model, which may be attributed to the discrepancy in the capacity gaps between the teacher model (LaBSE) and the student model (``small'' or ``base'').\footnote{$\mathcal{L}_{ld}$ can hardly work for UN and BUCC as they contain hundreds of thousands of candidates for the model to score, which is more complicated than the 1,000 candidates of Tatoeba.} Refer to Appx.~\ref{sec:app-ablation} for all detailed results in this section.

\section{Conclusion}
We presented LEALLA, a lightweight model for generating low-dimensional language-agnostic sentence embeddings. Experimental results showed that LEALLA could yield solid performance for 109 languages after distilling knowledge from LaBSE. Future work can focus on reducing the vocabulary size of LaBSE to shrink the model further and exploring the effectiveness of lightweight model pre-training for parallel sentence alignment.

\section*{Limitations}
In this study, we used the same training data as LaBSE (refer to Fig. 7 of~\citep{feng-etal-2022-language}), where more training data for high-resource languages may cause the biased model accuracy for those languages. Second, evaluation for low-resource languages in this study depended only on the Tatoeba benchmark, which contains only 1,000 positive sentence pairs for each language with English. The same limitation exists in all the related work, such as LaBSE and LASER. Further evaluation for low-resource languages will be necessary in the future once larger evaluation benchmarks, including over 100k gold parallel sentences for low-resource languages, are available. Third, all the training data used in this work are English-centric sentence pairs, which may result in the inferior model performance for aligning parallel sentences between non-English language pairs.



\section*{Acknowledgements}
We would like to thank our colleagues from Translate, Descartes, and other Google teams for their valuable contributions and feedback. A special mention to Fangxiaoyu Feng, Shuying Zhang, Gustavo Hernandez Abrego, and Jianmon Ni for their support in sharing information on LaBSE, and providing training data, expertise on language-agnostic sentence embeddings, and assistance with evaluation. We would also like to thank the reviewers for their insightful comments for improving the paper.

\bibliography{anthology,custom}

\begin{thebibliography}{20}
\expandafter\ifx\csname natexlab\endcsname\relax\def\natexlab#1{#1}\fi

\bibitem[{Artetxe and Schwenk(2019{\natexlab{a}})}]{artetxe-schwenk-2019-margin}
Mikel Artetxe and Holger Schwenk. 2019{\natexlab{a}}.
\newblock \href {https://doi.org/10.18653/v1/P19-1309} {Margin-based parallel corpus mining with multilingual sentence embeddings}.
\newblock In \emph{Proceedings of the 57th Annual Meeting of the Association for Computational Linguistics}, pages 3197--3203, Florence, Italy. Association for Computational Linguistics.

\bibitem[{Artetxe and Schwenk(2019{\natexlab{b}})}]{artetxe-schwenk-2019-massively}
Mikel Artetxe and Holger Schwenk. 2019{\natexlab{b}}.
\newblock \href {https://doi.org/10.1162/tacl_a_00288} {Massively multilingual sentence embeddings for zero-shot cross-lingual transfer and beyond}.
\newblock \emph{Transactions of the Association for Computational Linguistics}, 7:597--610.

\bibitem[{Conneau et~al.(2020)Conneau, Khandelwal, Goyal, Chaudhary, Wenzek, Guzm{\'a}n, Grave, Ott, Zettlemoyer, and Stoyanov}]{conneau-etal-2020-unsupervised}
Alexis Conneau, Kartikay Khandelwal, Naman Goyal, Vishrav Chaudhary, Guillaume Wenzek, Francisco Guzm{\'a}n, Edouard Grave, Myle Ott, Luke Zettlemoyer, and Veselin Stoyanov. 2020.
\newblock \href {https://doi.org/10.18653/v1/2020.acl-main.747} {Unsupervised cross-lingual representation learning at scale}.
\newblock In \emph{Proceedings of the 58th Annual Meeting of the Association for Computational Linguistics}, pages 8440--8451, Online. Association for Computational Linguistics.

\bibitem[{Feng et~al.(2022)Feng, Yang, Cer, Arivazhagan, and Wang}]{feng-etal-2022-language}
Fangxiaoyu Feng, Yinfei Yang, Daniel Cer, Naveen Arivazhagan, and Wei Wang. 2022.
\newblock \href {https://doi.org/10.18653/v1/2022.acl-long.62} {Language-agnostic {BERT} sentence embedding}.
\newblock In \emph{Proceedings of the 60th Annual Meeting of the Association for Computational Linguistics (Volume 1: Long Papers)}, pages 878--891, Dublin, Ireland. Association for Computational Linguistics.

\bibitem[{Guo et~al.(2018)Guo, Shen, Yang, Ge, Cer, Hernandez~Abrego, Stevens, Constant, Sung, Strope, and Kurzweil}]{guo-etal-2018-effective}
Mandy Guo, Qinlan Shen, Yinfei Yang, Heming Ge, Daniel Cer, Gustavo Hernandez~Abrego, Keith Stevens, Noah Constant, Yun-Hsuan Sung, Brian Strope, and Ray Kurzweil. 2018.
\newblock \href {https://doi.org/10.18653/v1/W18-6317} {Effective parallel corpus mining using bilingual sentence embeddings}.
\newblock In \emph{Proceedings of the Third Conference on Machine Translation: Research Papers}, pages 165--176, Brussels, Belgium. Association for Computational Linguistics.

\bibitem[{Hinton et~al.(2015)Hinton, Vinyals, and Dean}]{DBLP:journals/corr/HintonVD15}
Geoffrey~E. Hinton, Oriol Vinyals, and Jeffrey Dean. 2015.
\newblock \href {http://arxiv.org/abs/1503.02531} {Distilling the knowledge in a neural network}.
\newblock \emph{CoRR}, abs/1503.02531.

\bibitem[{Loshchilov and Hutter(2019)}]{DBLP:conf/iclr/LoshchilovH19}
Ilya Loshchilov and Frank Hutter. 2019.
\newblock \href {https://openreview.net/forum?id=Bkg6RiCqY7} {Decoupled weight decay regularization}.
\newblock In \emph{7th International Conference on Learning Representations, {ICLR} 2019, New Orleans, LA, USA, May 6-9, 2019}. OpenReview.net.

\bibitem[{Mao et~al.(2022)Mao, Chu, and Kurohashi}]{DBLP:journals/corr/abs-2205-15744}
Zhuoyuan Mao, Chenhui Chu, and Sadao Kurohashi. 2022.
\newblock \href {https://doi.org/10.48550/arXiv.2205.15744} {{EMS:} efficient and effective massively multilingual sentence representation learning}.
\newblock \emph{CoRR}, abs/2205.15744.

\bibitem[{Mao et~al.(2021)Mao, Gupta, Chu, Jaggi, and Kurohashi}]{mao-etal-2021-lightweight}
Zhuoyuan Mao, Prakhar Gupta, Chenhui Chu, Martin Jaggi, and Sadao Kurohashi. 2021.
\newblock \href {https://doi.org/10.18653/v1/2021.acl-long.226} {Lightweight cross-lingual sentence representation learning}.
\newblock In \emph{Proceedings of the 59th Annual Meeting of the Association for Computational Linguistics and the 11th International Joint Conference on Natural Language Processing (Volume 1: Long Papers)}, pages 2902--2913, Online. Association for Computational Linguistics.

\bibitem[{Pierre~Zweigenbaum and Rapp(2018)}]{ZWEIGENBAUM18.12}
Serge~Sharoff Pierre~Zweigenbaum and Reinhard Rapp. 2018.
\newblock \href {http://lrec-conf.org/workshops/lrec2018/W8/pdf/12_W8.pdf} {Overview of the third bucc shared task: Spotting parallel sentences in comparable corpora}.
\newblock In \emph{Proceedings of the Eleventh International Conference on Language Resources and Evaluation (LREC 2018)}, Paris, France. European Language Resources Association (ELRA).

\bibitem[{Reimers and Gurevych(2020)}]{reimers-gurevych-2020-making}
Nils Reimers and Iryna Gurevych. 2020.
\newblock \href {https://doi.org/10.18653/v1/2020.emnlp-main.365} {Making monolingual sentence embeddings multilingual using knowledge distillation}.
\newblock In \emph{Proceedings of the 2020 Conference on Empirical Methods in Natural Language Processing (EMNLP)}, pages 4512--4525, Online. Association for Computational Linguistics.

\bibitem[{Romero et~al.(2015)Romero, Ballas, Kahou, Chassang, Gatta, and Bengio}]{DBLP:journals/corr/RomeroBKCGB14}
Adriana Romero, Nicolas Ballas, Samira~Ebrahimi Kahou, Antoine Chassang, Carlo Gatta, and Yoshua Bengio. 2015.
\newblock \href {http://arxiv.org/abs/1412.6550} {Fitnets: Hints for thin deep nets}.
\newblock In \emph{3rd International Conference on Learning Representations, {ICLR} 2015, San Diego, CA, USA, May 7-9, 2015, Conference Track Proceedings}.

\bibitem[{Schwenk et~al.(2021)Schwenk, Chaudhary, Sun, Gong, and Guzm{\'a}n}]{schwenk-etal-2021-wikimatrix}
Holger Schwenk, Vishrav Chaudhary, Shuo Sun, Hongyu Gong, and Francisco Guzm{\'a}n. 2021.
\newblock \href {https://doi.org/10.18653/v1/2021.eacl-main.115} {{W}iki{M}atrix: Mining 135{M} parallel sentences in 1620 language pairs from {W}ikipedia}.
\newblock In \emph{Proceedings of the 16th Conference of the European Chapter of the Association for Computational Linguistics: Main Volume}, pages 1351--1361, Online. Association for Computational Linguistics.

\bibitem[{Sennrich et~al.(2016)Sennrich, Haddow, and Birch}]{sennrich-etal-2016-neural}
Rico Sennrich, Barry Haddow, and Alexandra Birch. 2016.
\newblock \href {https://doi.org/10.18653/v1/P16-1162} {Neural machine translation of rare words with subword units}.
\newblock In \emph{Proceedings of the 54th Annual Meeting of the Association for Computational Linguistics (Volume 1: Long Papers)}, pages 1715--1725, Berlin, Germany. Association for Computational Linguistics.

\bibitem[{Sun et~al.(2020)Sun, Yu, Song, Liu, Yang, and Zhou}]{sun-etal-2020-mobilebert}
Zhiqing Sun, Hongkun Yu, Xiaodan Song, Renjie Liu, Yiming Yang, and Denny Zhou. 2020.
\newblock \href {https://doi.org/10.18653/v1/2020.acl-main.195} {{M}obile{BERT}: a compact task-agnostic {BERT} for resource-limited devices}.
\newblock In \emph{Proceedings of the 58th Annual Meeting of the Association for Computational Linguistics}, pages 2158--2170, Online. Association for Computational Linguistics.

\bibitem[{Tang et~al.(2019)Tang, Lu, Liu, Mou, Vechtomova, and Lin}]{DBLP:journals/corr/abs-1903-12136}
Raphael Tang, Yao Lu, Linqing Liu, Lili Mou, Olga Vechtomova, and Jimmy Lin. 2019.
\newblock \href {http://arxiv.org/abs/1903.12136} {Distilling task-specific knowledge from {BERT} into simple neural networks}.
\newblock \emph{CoRR}, abs/1903.12136.

\bibitem[{Yang et~al.(2019)Yang, {\'{A}}brego, Yuan, Guo, Shen, Cer, Sung, Strope, and Kurzweil}]{DBLP:conf/ijcai/YangAYGSCSSK19}
Yinfei Yang, Gustavo~Hern{\'{a}}ndez {\'{A}}brego, Steve Yuan, Mandy Guo, Qinlan Shen, Daniel Cer, Yun{-}Hsuan Sung, Brian Strope, and Ray Kurzweil. 2019.
\newblock \href {https://doi.org/10.24963/ijcai.2019/746} {Improving multilingual sentence embedding using bi-directional dual encoder with additive margin softmax}.
\newblock In \emph{Proceedings of the Twenty-Eighth International Joint Conference on Artificial Intelligence, {IJCAI} 2019, Macao, China, August 10-16, 2019}, pages 5370--5378. ijcai.org.

\bibitem[{Yang et~al.(2020)Yang, Cer, Ahmad, Guo, Law, Constant, Hernandez~Abrego, Yuan, Tar, Sung, Strope, and Kurzweil}]{yang-etal-2020-multilingual}
Yinfei Yang, Daniel Cer, Amin Ahmad, Mandy Guo, Jax Law, Noah Constant, Gustavo Hernandez~Abrego, Steve Yuan, Chris Tar, Yun-hsuan Sung, Brian Strope, and Ray Kurzweil. 2020.
\newblock \href {https://doi.org/10.18653/v1/2020.acl-demos.12} {Multilingual universal sentence encoder for semantic retrieval}.
\newblock In \emph{Proceedings of the 58th Annual Meeting of the Association for Computational Linguistics: System Demonstrations}, pages 87--94, Online. Association for Computational Linguistics.

\bibitem[{Yim et~al.(2017)Yim, Joo, Bae, and Kim}]{DBLP:conf/cvpr/YimJBK17}
Junho Yim, Donggyu Joo, Ji{-}Hoon Bae, and Junmo Kim. 2017.
\newblock \href {https://doi.org/10.1109/CVPR.2017.754} {A gift from knowledge distillation: Fast optimization, network minimization and transfer learning}.
\newblock In \emph{2017 {IEEE} Conference on Computer Vision and Pattern Recognition, {CVPR} 2017, Honolulu, HI, USA, July 21-26, 2017}, pages 7130--7138. {IEEE} Computer Society.

\bibitem[{Ziemski et~al.(2016)Ziemski, Junczys-Dowmunt, and Pouliquen}]{ziemski-etal-2016-united}
Micha{\l} Ziemski, Marcin Junczys-Dowmunt, and Bruno Pouliquen. 2016.
\newblock \href {https://aclanthology.org/L16-1561} {The {U}nited {N}ations parallel corpus v1.0}.
\newblock In \emph{Proceedings of the Tenth International Conference on Language Resources and Evaluation ({LREC}'16)}, pages 3530--3534, Portoro{\v{z}}, Slovenia. European Language Resources Association (ELRA).

\end{thebibliography}
\bibliographystyle{acl_natbib}

\appendix

\section{Evaluation Benchmarks}
\label{sec:app-eval}
Tatoeba~\citep{artetxe-schwenk-2019-massively} supports the evaluation across 112 languages and contains up to 1,000 sentence pairs for each language and English. The languages of Tatoeba that are not included in the training data of LaBSE and LEALLA serve as the evaluation for unseen languages. UN~\citep{ziemski-etal-2016-united} is composed of 86,000 aligned bilingual documents for en-ar, en-es, en-fr, en-ru, and en-zh. Following~\citet{feng-etal-2022-language}, we evaluate the model performance for es, fr, ru, and zh on the UN task. There are about 9.5M sentence pairs for each language with English after deduping. BUCC shared task~\citep{ZWEIGENBAUM18.12} is a benchmark to mine parallel sentences from comparable corpora. We conduct the evaluation using BUCC2018 tasks for en-de, en-fr, en-ru, and en-zh, following the setting of~\citet{reimers-gurevych-2020-making}.\footnote{\url{https://github.com/UKPLab/sentence-transformers/blob/master/examples/applications/parallel-sentence-mining/bucc2018.py}} For the results of LaBSE reported in Table~\ref{tab:LEALLA}, we re-conduct the evaluation experiments using the open-sourced model of LaBSE.\footnote{\url{https://tfhub.dev/google/LaBSE}}

\begin{table*}[t]
    \centering
    \resizebox{0.8\linewidth}{!}{
    \begin{tabular}{l|r|rrrrr|rrrrr}
        \toprule
        \multirow{2}{*}{\textbf{Model}} & \multirow{2}{*}{\textbf{Tatoeba}} & \multicolumn{5}{c|}{\textbf{UN}} & \multicolumn{5}{c}{\textbf{BUCC}} \\
        & & \textbf{es} & \textbf{fr} & \textbf{ru} & \textbf{zh} & \textbf{avg.} & \textbf{de} & \textbf{fr} & \textbf{ru} & \textbf{zh} & \textbf{avg.} \\
        \toprule
        \multicolumn{9}{l}{\textbf{LEALLA-small}} \\
        $\mathcal{L}_{ams}$ & 80.3 & 88.1 & 85.2 & 88.0 & 83.9 & 86.3 & 93.0 & 89.7 & 90.6 & 88.3 & 90.4 \\
        $\mathcal{L}_{ams}+\mathcal{L}_{fd}$ & \textbf{80.6} & 89.3 & \textbf{86.8} & 88.0 & \textbf{84.0} & \textbf{87.0} & \textbf{93.9} & \textbf{90.6} & \textbf{91.4} & \textbf{89.7} & \textbf{91.4} \\
        $\mathcal{L}_{ams}+\mathcal{L}_{df}$ & 80.0 & \textbf{89.4} & 86.3 & \textbf{88.1} & 83.9 & 86.9 & 93.8 & 90.1 & 91.1 & 88.9 & 91.0 \\
        $\mathcal{L}_{ams}+\mathcal{L}_{syn}$ & 80.2 & 88.5 & 85.0 & 87.1 & 82.8 & 85.9 & 93.6 & 89.9 & 90.9 & 88.7 & 90.8 \\
        \hline
        \multicolumn{9}{l}{\textbf{LEALLA-base}} \\
        $\mathcal{L}_{ams}$ & 81.7 & 89.8 & 85.9 & 88.6 & 85.4 & 87.4 & 94.2 & 91.0 & 91.3 & \textbf{91.1} & 91.9 \\
        $\mathcal{L}_{ams}+\mathcal{L}_{fd}$ & \textbf{82.2} & \textbf{90.2} & \textbf{87.5} & \textbf{89.4} & \textbf{86.8} & \textbf{88.5} & \textbf{95.0} & \textbf{91.6} & \textbf{91.7} & 91.0 & \textbf{92.3} \\
        $\mathcal{L}_{ams}+\mathcal{L}_{df}$ & 81.8 & 90.0 & 87.3 & 89.2 & 86.3 & 88.2 & 94.7 & 91.4 & \textbf{91.7} & 90.9 & 92.2 \\
        $\mathcal{L}_{ams}+\mathcal{L}_{syn}$ & 81.9 & 89.7 & 86.7 & 88.8 & 85.9 & 87.8 & 94.5 & 91.1 & \textbf{91.7} & 90.3 & 91.9 \\
        \hline
        \multicolumn{9}{l}{\textbf{LEALLA-large}} \\
        $\mathcal{L}_{ams}$ & 82.9 & 90.1 & 87.1 & 89.3 & 87.4 & 88.5 & 94.6 & 91.2 & 91.5 & 91.4 & 92.2 \\
        $\mathcal{L}_{ams}+\mathcal{L}_{fd}$ & \textbf{83.4} & \textbf{90.6} & \textbf{88.4} & \textbf{89.8} & \textbf{87.7} & \textbf{89.1} & \textbf{95.3} & \textbf{92.0} & \textbf{92.0} & \textbf{92.0} & \textbf{92.8} \\
        $\mathcal{L}_{ams}+\mathcal{L}_{df}$ & 83.0 & 90.3 & 87.6 & 89.7 & 87.2 & 88.7 & \textbf{95.3} & 91.9 & \textbf{92.0} & 91.7 & 92.7 \\
        $\mathcal{L}_{ams}+\mathcal{L}_{syn}$ & 83.0 & 90.0 & 87.4 & 89.7 & 86.8 & 88.5 & 94.9 & 91.7 & 91.8 & 91.4 & 92.5 \\
        \bottomrule
    \end{tabular}
    }
    \caption{Results of comparisons among three feature distillation objectives. $\mathcal{L}_{df}$ and $\mathcal{L}_{syn}$ indicate ``\textit{Distillation-first}'' and ``\textit{Synchronized}'' objectives in Fig.~\ref{fig:feature-distill}.}
    \label{tab:feature-distill}
\end{table*}

\begin{table}[t]
    \centering
    \resizebox{0.8\linewidth}{!}{
    \begin{tabular}{lr}
        \toprule
        Hyperparameter & Bound \\
        \toprule
        $\alpha$ & 1 \\
        $\beta$ & 1e02, 1e03, 1e04, 1e05 \\
        $\gamma$ & 1e-01, 1e-02, 1e-03 \\
        batch size & 2,048, 4,096, 8,192 \\
        learning rate & 1e-4, 5e-4, 1e-3 \\
        \bottomrule
    \end{tabular}
    }
    \caption{Hyperparameter bounds.}
    \label{tab:hparams}
\end{table}

\section{Training Details}
\label{sec:app-training}

All of the models in this work are trained with the same training data and development data as LaBSE~\citep{feng-etal-2022-language}. Refer to Section 3.1 and Appx. C of \citet{feng-etal-2022-language} for dataset and supported language details. We train models on Cloud TPU V3 with 32-cores with a global batch size of 8,192 sentences and a maximum sequence length of 128. For a fair comparison with LaBSE for more than 109 languages, we use the 501k vocabulary of LaBSE (trained with BPE~\citep{sennrich-etal-2016-neural}) and do not consider modifying its size in this work. We employ AdamW~\citep{DBLP:conf/iclr/LoshchilovH19} for optimizing the model using the initial learning rate of 1e-03 for models with a hidden state size larger than 384 and 5e-04 for models with a hidden state size smaller than 256. For LEALLA-small and LEALLA-base, $\alpha$, $\beta$, and $\gamma$ are set as 1, 1e03 and 1e-02. For LEALLA-large, they are set as 1, 1e04, and 1e-02, respectively. $T$ in Eq.~\ref{eq:ld} is set to 100. All the models in Section~\ref{sec:labse-rd} are trained for 5k steps. Models in Section~\ref{sec:lite-arch} and Section~\ref{sec:kd} with a hidden state size over 256 are trained for 200k steps, and those with a hidden state size below 192 are trained for 100k steps. It costs around 24 hours, 36 hours, and 48 hours to train LEALLA-small, LEALLA-base, and LEALLA-large, respectively. Hyperparameters are tuned using a held-out development dataset following~\citet{feng-etal-2022-language} with a grid search. The bounds tuned for each hyperparameter are shown in Table~\ref{tab:hparams}.

\begin{table*}[t]
    \centering
    \resizebox{0.8\linewidth}{!}{
    \begin{tabular}{l|r|rrrrr|rrrrr}
        \toprule
        \multirow{2}{*}{\textbf{Dimension}} & \multirow{2}{*}{\textbf{Tatoeba}} & \multicolumn{5}{c|}{\textbf{UN}} & \multicolumn{5}{c}{\textbf{BUCC}} \\
        & & \textbf{es} & \textbf{fr} & \textbf{ru} & \textbf{zh} & \textbf{avg.} & \textbf{de} & \textbf{fr} & \textbf{ru} & \textbf{zh} & \textbf{avg.} \\
        \toprule
        768 (LaBSE) & 83.7 & 90.8 & 89.0 & 90.4 & 88.3 & 89.6 & 95.5 & 92.3 & 92.2 & 92.5 & 93.1 \\
        512 & 83.7 & 90.1 & 88.1 & 89.7 & 87.4 & 88.8 & 95.4 & 92.1 & 92.0 & 92.4 & 93.0 \\
        384 & 83.7 & 90.1 & 88.1 & 89.6 & 87.4 & 88.8 & 95.5 & 92.0 & 92.0 & 92.6 & 93.0 \\
        256 & 83.6 & 90.3 & 87.9 & 89.2 & 87.4 & 88.7 & 95.3 & 92.0 & 92.1 & 92.2 & 92.9 \\
        192 & 83.4 & 89.8 & 87.5 & 89.5 & 87.0 & 88.5 & 95.2 & 91.9 & 91.9 & 92.2 & 92.8 \\
        128 & 83.1 & 89.2 & 86.9 & 88.6 & 85.9 & 87.7 & 95.1 & 91.4 & 91.8 & 91.6 & 92.5 \\
        64 & 81.8 & 88.4 & 84.4 & 87.3 & 83.8 & 86.0 & 93.9 & 89.8 & 90.7 & 88.9 & 90.8 \\
        32 & 78.4 & 82.7 & 74.8 & 80.4 & 73.7 & 77.9 & 87.1 & 81.5 & 84.1 & 75.5 & 82.1 \\
        \bottomrule
    \end{tabular}
    }
    \caption{Results of the dimension-reduced LaBSE embeddings.}
    \label{tab:labse-to-x}
\end{table*}

\begin{table*}[t]
    \centering
    \resizebox{\linewidth}{!}{
    \begin{tabular}{c|rrrrrr|r|rrrrr|rrrrr}
        \toprule
        \multirow{2}{*}{\textbf{\#}} & \multirow{2}{*}{$\mathbf{L}$} & \multirow{2}{*}{$\mathbf{d_h}$} & \multirow{2}{*}{$\mathbf{d_{ff}}$} & \multirow{2}{*}{$\mathbf{H}$} & \multirow{2}{*}{$\mathbf{P}$} & \multirow{2}{*}{$\mathbf{P_E}$} & \multirow{2}{*}{\textbf{Tatoeba}} & \multicolumn{5}{c|}{\textbf{UN}} & \multicolumn{5}{c}{\textbf{BUCC}} \\
        & & & & & & & & \textbf{es} & \textbf{fr} & \textbf{ru} & \textbf{zh} & \textbf{avg.} & \textbf{de} & \textbf{fr} & \textbf{ru} & \textbf{zh} & \textbf{avg.} \\
        \toprule
        \multicolumn{9}{l}{\textbf{LaBSE}} \\
        0 & 12 & 768 & 3072 & 12 & 471M & 85M & 83.7 & 90.8 & 89.0 & 90.4 & 88.3 & 89.6 & 95.5 & 92.3 & 92.2 & 92.5 & 93.1 \\
        \hline
        \multicolumn{9}{l}{\textbf{Fewer Layers}} \\
        1 & 6 & 768 & 3072 & 12 & 428M & 42M & 82.9 & 90.2 & 87.4 & 89.2 & 87.4 & 88.6 & 94.3 & 90.9 & 91.2 & 91.1 & 91.9 \\
        2 & 3 & 768 & 3072 & 12 & 407M & 21M & 82.2 & 89.4 & 86.1 & 88.0 & 86.5 & 87.5 & 93.7 & 90.1 & 90.8 & 90.1 & 91.2 \\
        \hline
        \multicolumn{9}{l}{\textbf{Smaller Hidden Size}} \\
        3 & 12 & 384 & 1536 & 12 & 214M & 21M & 82.6 & 90.1 & 86.9 & 89.6 & 87.0 & 88.4 & 94.4 & 91.2 & 91.4 & 91.3 & 92.1 \\
        4 & 12 & 192 & 768 & 12 & 102M & 6M & 81.0 & 89.4 & 85.6 & 88.1 & 85.0 & 87.0 & 93.6 & 90.4 & 91.1 & 89.9 & 91.3 \\
        \hline
        \multicolumn{9}{l}{\textbf{Thin-deep Architecture}} \\
        5 & 24 & 384 & 1536 & 12 & 235M & 42M & 83.2 & 90.6 & 87.3 & 89.2 & 87.4 & 88.6 & 94.7 & 91.5 & 91.6 & 91.9 & 92.4 \\
        6 & 24 & 256 & 1024 & 8 & 147M & 19M & 82.9 & 90.1 & 87.1 & 89.3 & 87.4 & 88.5 & 94.6 & 91.2 & 91.5 & 91.4 & 92.2 \\
        7 & 24 & 192 & 768 & 12 & 107M & 11M & 81.7 & 89.8 & 85.9 & 88.6 & 85.4 & 87.4 & 94.2 & 91.0 & 91.3 & 91.1 & 91.9 \\
        8 & 24 & 128 & 512 & 8 & 69M & 5M & 80.3 & 88.1 & 85.2 & 88.0 & 83.9 & 86.3 & 93.0 & 89.7 & 90.6 & 88.3 & 90.4 \\
        9 & 24 & 64 & 256 & 8 & 33M & 1M & 75.2 & 83.7 & 78.6 & 83.0 & 72.1 & 79.4 & 87.9 & 83.0 & 86.0 & 75.1 & 83.0 \\
        \hline
        \multicolumn{9}{l}{\textbf{MobileBERT-like Thin-deep Architecture}} \\
        10 & 24 & 256 & 256 & 4 & 138M & 10M & 82.1 & 89.4 & 86.5 & 88.4 & 86.5 & 87.7 & 94.1 & 91.0 & 91.0 & 91.7 & 92.0 \\
        11 & 24 & 192 & 192 & 4 & 102M & 6M & 81.0 & 89.0 & 85.4 & 88.5 & 85.3 & 87.1 & 93.8 & 90.3 & 91.0 & 89.9 & 91.3 \\
        12 & 24 & 128 & 128 & 4 & 66M & 2M & 79.7 & 88.1 & 84.1 & 87.6 & 83.3 & 85.8 & 92.6 & 88.8 & 90.4 & 87.6 & 89.9 \\
        \bottomrule
    \end{tabular}
    }
    \caption{Results of thin-deep and MobileBERT-like architectures. $\mathbf{L}$, $\mathbf{d_h}$, $\mathbf{d_{ff}}$, $\mathbf{H}$, $\mathbf{P}$, and $\mathbf{P_E}$ indicate the number of layers, dimension of hidden states, dimension of feed-forward hidden states, number of attention heads, number of model parameters, and number of encoder parameters (except for the word embedding layer).}
    \label{tab:mobile}
\end{table*}

\begin{table*}[t!]
    \centering
    \resizebox{0.85\linewidth}{!}{
    \begin{tabular}{l|r|rrrrr|rrrrr}
        \toprule
        \multirow{2}{*}{\textbf{Model}} & \multirow{2}{*}{\textbf{Tatoeba}} & \multicolumn{5}{c|}{\textbf{UN}} & \multicolumn{5}{c}{\textbf{BUCC}} \\
        & & \textbf{es} & \textbf{fr} & \textbf{ru} & \textbf{zh} & \textbf{avg.} & \textbf{de} & \textbf{fr} & \textbf{ru} & \textbf{zh} & \textbf{avg.} \\
        \toprule
        \multicolumn{9}{l}{\textbf{LEALLA-small}} \\
        $\mathcal{L}_{ams}$ & 80.3 & 88.1 & 85.2 & 88.0 & 83.9 & 86.3 & 93.0 & 89.7 & 90.6 & 88.3 & 90.4 \\
        $\mathcal{L}_{fd}$ & 78.2 & 89.0 & 84.6 & 87.5 & 79.6 & 85.2 & \textbf{94.2} & 90.5 & 91.2 & 88.9 & 91.2 \\
        $\mathcal{L}_{ld}$ & 75.1 & 1.5 & 1.1 & 0.9 & 5.6 & 2.3 & 0.1 & 0.0 & 0.1 & 0.0 & 0.1 \\
        $\mathcal{L}_{ams}+\mathcal{L}_{fd}$ & 80.6 & 89.3 & \textbf{86.8} & 88.0 & 84.0 & 87.0 & 93.9 & \textbf{90.6} & \textbf{91.4} & 89.7 & 91.4 \\
        $\mathcal{L}_{ams}+\mathcal{L}_{ld}$ & 80.6 & \textbf{89.6} & 85.8 & 88.6 & 84.4 & 87.1 & 94.1 & 90.3 & 91.2 & 90.0 & 91.4 \\
        $\mathcal{L}_{ams}+\mathcal{L}_{fd}+\mathcal{L}_{ld}$ & \textbf{80.7} & 89.4 & 86.0 & \textbf{88.7} & \textbf{84.9} & \textbf{87.3} & 94.0 & \textbf{90.6} & 91.2 & \textbf{90.3} & \textbf{91.5} \\
        \hline
        \multicolumn{9}{l}{\textbf{LEALLA-base}} \\
        $\mathcal{L}_{ams}$ & 81.7 & 89.8 & 85.9 & 88.6 & 85.4 & 87.4 & 94.2 & 91.0 & 91.3 & 91.1 & 91.9 \\
        $\mathcal{L}_{fd}$ & 81.1 & 90.2 & 87.3 & 89.4 & 85.5 & 88.1 & \textbf{95.0} & \textbf{91.6} & \textbf{91.8} & 91.3 & \textbf{92.4} \\
        $\mathcal{L}_{ld}$ & 80.6 & 66.3 & 49.4 & 51.0 & 85.7 & 63.1 & 57.5 & 80.1 & 60.6 & 88.6 & 71.7 \\
        $\mathcal{L}_{ams}+\mathcal{L}_{fd}$ & 82.2 & 90.2 & \textbf{87.5} & 89.4 & 86.8 & 88.5 & \textbf{95.0} & \textbf{91.6} & 91.7 & 91.0 & 92.3 \\
        $\mathcal{L}_{ams}+\mathcal{L}_{ld}$ & 82.3 & 90.0 & \textbf{87.5} & 89.2 & 86.8 & 88.4 & 94.8 & 91.3 & 91.6 & \textbf{91.4} & 92.3 \\
        $\mathcal{L}_{ams}+\mathcal{L}_{fd}+\mathcal{L}_{ld}$ & \textbf{82.4} & \textbf{90.3} & 87.4 & \textbf{89.8} & \textbf{87.2} & \textbf{88.7} & 94.9 & 91.4 & \textbf{91.8} & \textbf{91.4} & \textbf{92.4} \\
        \hline
        \multicolumn{9}{l}{\textbf{LEALLA-large}} \\
        $\mathcal{L}_{ams}$ & 82.9 & 90.1 & 87.1 & 89.3 & 87.4 & 88.5 & 94.6 & 91.2 & 91.5 & 91.4 & 92.2 \\
        $\mathcal{L}_{fd}$ & 82.4 & 89.8 & 87.2 & 89.4 & 86.1 & 88.1 & \textbf{95.3} & 91.8 & 92.0 & \textbf{92.2} & \textbf{92.8} \\
        $\mathcal{L}_{ld}$ & 82.3 & 87.2 & 78.8 & 83.3 & 86.9 & 84.1 & 88.4 & 87.4 & 86.9 & 91.8 & 88.6 \\
        $\mathcal{L}_{ams}+\mathcal{L}_{fd}$ & 83.4 & 90.6 & 88.4 & 89.8 & 87.7 & 89.1 & \textbf{95.3} & \textbf{92.0} & 92.0 & 92.0 & \textbf{92.8} \\
        $\mathcal{L}_{ams}+\mathcal{L}_{ld}$ & 83.4 & 90.6 & 87.9 & \textbf{90.0} & 87.7 & 89.1 & \textbf{95.3} & 91.8 & 91.7 & 92.4 & \textbf{92.8} \\
        $\mathcal{L}_{ams}+\mathcal{L}_{fd}+\mathcal{L}_{ld}$ & \textbf{83.5} & \textbf{90.8} & \textbf{88.5} & 89.9 & \textbf{87.9} & \textbf{89.3} & \textbf{95.3} & \textbf{92.0} & \textbf{92.1} & 91.9 & \textbf{92.8} \\
        \bottomrule
    \end{tabular}
    }
    \caption{Results of LEALLA with different loss functions and loss combinations.}
    \label{tab:ablation-all}
\end{table*}

\section{Discussion about Feature Distillation}
\label{app:distillation}

\begin{figure}[t]
    \centering
    \includegraphics[width=\linewidth]{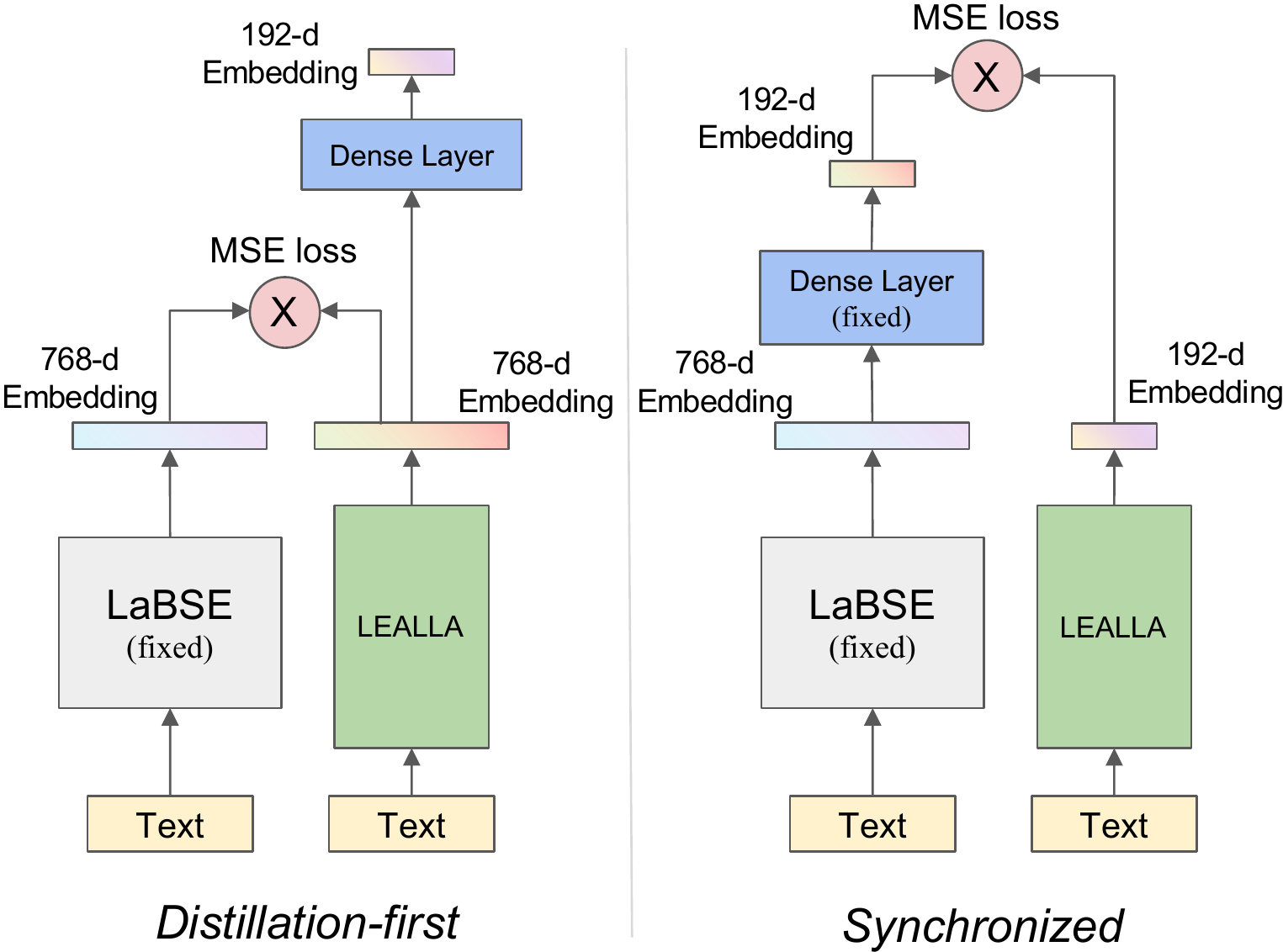}
    \caption{Another two patterns of feature distillation.}
    \label{fig:feature-distill}
\end{figure}

We additionally investigate another two patterns for feature distillation. As illustrated in Fig.~\ref{fig:feature-distill}, ``\textit{Distillation-first}'' modifies the position for computing the MSE loss compared with $\mathcal{L}_{fd}$ of Eq.~\ref{eq:fd}. The \texttt{[CLS]} pooler within the LEALLA encoder is used to generate 768-d embeddings first. A dense layer is employed to transform the 768-d embeddings to low-dimension after calculating the MSE loss. ``\textit{Synchronized}'' transforms the LaBSE embeddings to low-dimension, then the MSE loss is constructed between two low-dimensional embeddings. As the MSE loss is computed simultaneously with the AMS loss, it is denoted as ``\textit{Synchronized}''. For ``\textit{Synchronized}'', it requires a fixed dense layer to conduct the dimension reduction for the LaBSE embeddings, for which we utilize the pre-trained model introduced in Section~\ref{sec:labse-rd}. We denote these two patterns of feature distillation as $\mathcal{L}_{df}$ and $\mathcal{L}_{syn}$.

As reported in Table~\ref{tab:feature-distill}, $\mathcal{L}_{ams}+\mathcal{L}_{fd}$ ($\mathcal{L}_{fd}$ is feature distillation introduced in the main text) consistently outperforms $\mathcal{L}_{ams}+\mathcal{L}_{df}$ and $\mathcal{L}_{ams}+\mathcal{L}_{syn}$ in all the three LEALLA models. $\mathcal{L}_{ams}+\mathcal{L}_{df}$ and $\mathcal{L}_{ams}+\mathcal{L}_{syn}$ perform comparably on Tatoeba with the models trained without distillation loss. $\mathcal{L}_{ams}+\mathcal{L}_{df}$ obtains performance gains for high-resource languages on UN and BUCC compared with $\mathcal{L}_{ams}$, but still underperforms $\mathcal{L}_{ams}+\mathcal{L}_{fd}$.

$\mathcal{L}_{df}$ forces the lightweight model to approximate the teacher embeddings first in the intermediate part of the model, on top of which the low-dimensional sentence embeddings are generated for computing the AMS loss, while $\mathcal{L}_{fd}$ (Eq.~\ref{eq:fd}) is calculated after computing the AMS loss. As the AMS loss directly indicates the evaluation tasks, we suppose $\mathcal{L}_{fd}$ is a more flexible objective for feature distillation. In addition, $\mathcal{L}_{syn}$ is not beneficial because it depends on a dimension-reduced LaBSE, which is a less robust teacher compared with LaBSE.

\section{Results of Dimension-reduction Experiments}
\label{app:labse-to-x}

We report all the results of Section~\ref{sec:labse-rd} in Table~\ref{tab:labse-to-x}.

\section{Results of Thin-deep and MobileBERT-like Architectures}
\label{sec:app-mobile}

Table~\ref{tab:mobile} presents the detailed results of each architecture we explored in Section~\ref{sec:lite-arch}. Besides showing the results for each language on UN and BUCC for models \#0 - \#8, we provide the results of a further smaller thin-deep architecture (\#9) and MobileBERT-like~\citep{sun-etal-2020-mobilebert} thin-deep architectures (\#10 - \#12). The 64-d thin-deep architecture contains only 33M parameters. However, its performance on three evaluation benchmarks downgrades by up to 7.4 points compared with \#5 - \#8, which demonstrates that 128-d may be a lower bound as universal sentence embeddings for aligning parallel sentences for 109 languages. Moreover, \#10 - \#12 show the results of MobileBERT-like architectures whose feed-forward hidden size is identical to hidden size. They have fewer parameters than \#5 - \#8, but they perform worse than \#5 - \#8, respectively (e.g., compare \#10 with \#6). Therefore, we did not employ MobileBERT-like architectures for LEALLA.

\section{Results of Ablation Study}
\label{sec:app-ablation}

We report all the results of the ablation study (Section~\ref{sec:exps}) in Table~\ref{tab:ablation-all}.

\end{document}